\documentclass{article}
\usepackage{nips14submit_e,times}
\usepackage{amsmath,bm,algorithm,algorithmic}
\usepackage{amsfonts}
\usepackage{graphics,graphicx}
\usepackage{subfigure}
\author{Songlin Zhao\\
University of Florida\\
zhaosonglin@gmail.com
}
\title{Regularized Kernel Recursive Least Square Algoirthm}

\nipsfinalcopy 
\begin{document}
\maketitle
\begin{abstract}
In most adaptive signal processing applications, system linearity is assumed and adaptive linear filters are thus used. The traditional class of supervised adaptive filters rely on error-correction learning for their adaptive capability. The kernel method is a powerful nonparametric modeling tool for pattern analysis and statistical signal processing. Through a nonlinear mapping, kernel methods transform the data into a set of points in a Reproducing Kernel Hilbert Space. KRLS achieves high accuracy and has fast convergence rate in stationary scenario. However the good performance is obtained at a cost of high computation complexity. Sparsification in kernel methods is know to related to less computational complexity and memory consumption
\end{abstract}

\section{Linear And Nonlinear Adaptive Filter}
The term \emph{filter} usually refers to a system that is designed to extract information about a prescribed quantity of interest from noisy data. An \emph{adaptive filter} is a filter that self-adjusts its input output mapping according to an optimization algorithm like predictive coding~\cite{Huang11} mostly driven by an error signal. Because of the complexity of the optimization algorithms, most adaptive filters are digital filters. With the available processing capabilities of current digital signal processors, adaptive filters have become much more popular and are now widely used in various fields such as sound wave based communication devices~\cite{Huang15}, face extraction~\cite{wu2011learning,wu2014learning}, camcorders and digital cameras, information retrieval~\cite{Nan15} and medical monitoring equipments.

\subsection{Linear Adaptive Filter}
In most adaptive signal processing applications, system linearity is assumed and adaptive linear filters are thus used. The traditional class of supervised adaptive filters rely on error-correction learning for their adaptive capability. Therefore, the error is the necessary element of a cost function, which is a criterion for optimum performance of the filter. 

The linear filter includes a set of adaptively adjustable parameters (also known as weights), which is marked as $\bm \omega(n-1)$, where $n$ denotes discrete time. The input signal $\bm u(n)$ applied to the filter at time $n$, produces the actual response $y(n)$ via
\begin{equation}
y(n) = \bm \omega(n-1)^T \bm u(n)  \nonumber
\end{equation}
Then this actual response is compared with the corresponding desired response $d(n)$ to produce the error signal $e(n)$. The error signal, in turn, acts as a guide to adjust the weights $\bm \omega(n-1)$ by an incremental value denoted by $\triangle \bm \omega(n)$. On the next iteration, $\bm \omega(n)$ becomes the latest value of the weights to be updated. The adaptive filtering process is repeated continuously until the filter reaches a stop condition, which normally is that the weights adjustment is small enough.

An important issue in the adaptive design, no matter linear or nonlinear adaptive filter, is to ensure the learning curve is convergent with an increasing number of iterations. Under this condition, we define the system is in a steady-state.

\subsection{Nonlinear Adaptive Filter}
Even though linear adaptive filtering can approximate non-linearity, the performance of adaptive linear filters is not satisfactory in applications where nonlinearities are significant. Hence more advanced nonlinear models are required. At present, Neural Networks and \emph{Kernel Adaptive Filters} are popular nonlinear models.  By providing linearity in a high dimension feature space, Reproduce Kernel Hilbert Space (RKHS), and universal approximation in Euclidean space with universal kernels\cite{kernelbook}, the kernel adaptive filters are attracting more attention. Through a reproducing kernel, kernel adaptive filters map data from an input space to RKHS, where appropriate linear methods are applied to the transformed data. This procedure implements a nonlinear treatment for the data in the input space. Comparing with other nonlinear techniques, kernel adaptive filters have the following features:
\begin{itemize}
  \item They can be universal approximations whenever the kernel is universal.
  \item They have no local minima with the squared error cost function.
  \item They have moderate complexity in terms of computation and memory.
  \item They belong to online learning method~\cite{Huang14} and have good tracking ability to handle nonstationary conditions.
\end{itemize}
The details about kernel adaptive filters are introduced next.

\section{Reproducing Kernel Hilbert Space }
Reproducing Kernel Hilbert Space, RKHS for short, is a complete inner product space associate with a Mercer kernel.  A Mercer kernel is a continuous, symmetric and positive definite function $\kappa$ : $\mathbb U \times \mathbb U \rightarrow \mathbb R $, where $\mathbb U$ is the input domain in Euclidean space $\mathbb R^L$ ($L$ is the input order). The commonly used kernels includes the Gaussian kernel (\ref{equ:gaussianKernel}) and the polynomial kernel (\ref{equ:gpolyKernel})  \cite{kernelbook}.
\begin{equation}
\kappa(\bm u, \bm u^{\prime}) = \text{exp}(-\frac{\| \bm u - \bm u^{\prime} \|^2}{\sigma^2})
\label{equ:gaussianKernel}
\end{equation}
\begin{equation}
\label{equ:gpolyKernel}
\kappa(\bm u, \bm u^{\prime})= (\bm u^T \bm u^{\prime}+1)^p
\end{equation}
If only one free parameter is inserted into the kernel function, $\kappa(\bm u, \cdot)$ is expressed as a transformed feature vector  $\varphi(\bm u)$ through a mapping $\bm\varphi: \mathbb U \rightarrow \mathbb H $, where $\mathbb H$ is the RKHS. Therefore $\bm\varphi(\bm u) = \kappa(\bm u, \cdot) $.  One of the most important properties of RKHS for practical applications is the one, called the ``kernel trick''
\begin{equation}
\bm \varphi(\bm u)^T \bm \varphi(\bm u^{\prime})= \kappa(\bm u, \bm u^{\prime})
\end{equation}
that allows computing the inner products between two RKHS functions as a scalar evaluation in the input space by the kernel.

Besides ``kernel trick'', some other properties of RKHS related to this work are as follows. Assume $\mathbb H$ be any RKHS of all real-valued functions of $\bm u$ that are generated by the $\kappa(\bm u, \cdot)$. Suppose now two functions $h(\cdot)$ and $g(\cdot)$ are picked from the space $\mathbb H$ that are respectively represented by
\begin{equation}
h(\cdot) = \sum_{i=1}^l a_i\kappa(\bm c_i,\cdot)= \sum_{i=1}^l a_i \bm \varphi (\bm c_i) \nonumber
\end{equation}
\begin{equation}
g(\cdot) = \sum_{j=1}^m b_j\kappa(\tilde{\bm c}_j,\cdot)= \sum_{j=1}^m b_j \bm \varphi (\tilde{\bm c}_j)\nonumber
\end{equation}
where $a_i$ and $b_j$ are the expansion coefficients and both $\bm c_i$ and $\tilde{\bm c}_j \in \mathbb U$ for all $i$ and $j$.
\begin{enumerate}
  \item Symmetry
   \begin{equation}
   <h,g> = <g,h> \nonumber
   \end{equation}
  \item Scaling and distributive property
    \begin{equation}
    <(cf+dg),h> = c<f,h>+d<g,h> \nonumber
    \end{equation}
  \item Squared norm
    \begin{equation}
    \|h\|^2 = <h,h> \geq 0 \nonumber
    \end{equation}
\end{enumerate}
\section{Kernel Adaptive Filtering}
The kernel method is a powerful nonparametric modeling tool for pattern analysis and statistical signal processing. Through a nonlinear mapping, kernel methods transform the data into a set of points in a RKHS. Then various methods are utilized to find relationship between the data. There are many successful examples of this methodology including support vector machines (SVM) \cite{svm}, kernel regularization network \cite{krn}, kernel principal component analysis (kernel PCA) \cite{kcomponent} and kernel fisher discriminant analysis \cite{kernelfisher}. Kernel adaptive filters is a class of kernel methods. As a member of kernel adaptive filter, kernel affine projection algorithms (KAPA) include the kernel least mean square (KLMS) \cite{klms} as the simplest element and kernel recursive least squares (KRLS) \cite{KRLS} as the most computationally demanding.

The main idea of kernel adaptive filtering can be summarized as follows: Transform the input data into a high-dimension feature space $\mathbb F$, via a Mercer kernel. Then appropriate linear methods are subsequently applied on the transformed data. As long as a linear method in the feature space can be formulated in terms of inner products, there is no need to do computation in high-dimension space basing on ``kernel trick''. It has been proved that the kernel adaptive filters with universal kernel has universal approximation property. i.e. for any continuous input-output mapping $f: \mathbb{U}\rightarrow \mathbb{R}$, $\forall \varsigma >0, \exists \{ \bm u(i)\}_{i\in N} \in \mathbb{U}$ and real number ${c_i}_{i\in N}$, such that $\| f-\sum_i c_i\kappa(\bm u(i), \cdot)\|_2 < \varsigma$. This universal approximation property guarantees that the kernel method is capable of superior performance in nonlinear tasks. If we express a vector in $\mathbb F$ as
\begin{equation}
\bm \Omega=\sum_i c_i \bm \varphi(\bm u(i))
\end{equation}
we obtain
\begin{equation}
\|f-\Omega^T \bm \varphi\|_2 < \varsigma
\end{equation}

Furthermore, in the view of supervised learning, which requires the availability of a collection of desired responses, error cost functions (or error criteria) play significant role. Kernel adaptive filters provide a generalization of linear adaptive filters because these are a special case of the former when expressed in the dual space. Kernel adaptive filters exhibit a growing radial basis function network, learning the network topology and adaptive the free parameters directly from the data at the same time \cite{kernelbook}. In the following of this section, KRLS and KLMS are introduced respectively.

Consider the learning of a nonlinear function $f: \mathbb{U}\rightarrow \mathbb{R}$ based on a known sequence $(\bm{u}(1), d(1)), (\bm{u}(2), d(2)),\ldots, (\bm{u}(n), d(n))$, where $\mathbb{U}\in \mathbb{R}^L $ is the input space, $\bm{u}(i)$, $i= 1,...,n$ is the system input at sample time $i$, and $d(i)$ is the corresponding desired response.
\subsection{Kernel Recursive Least Square Algorithm}
The KRLS is actually the recursive least squares algorithm (RLS) algorithm in RKHS. At each iteration, one needs to solve the regularized least squares regression to obtain $f$:
\begin{equation}
\min_{f \in \mathbb{H}} \sum_{i=1}^n \|f(\bm u(i))-d(i)\|^2 + \lambda \|f\|_\mathbb{H}^2
\end{equation}
where $\lambda$ is the regularization term and $\|\cdot\|^2_\mathbb{H}$ denotes the norm in $\mathbb{H}$.

This problem can alternatively be solved in a feature space and results in KRLS. The learning problem of KRLS with regularization in feature space $\mathbb{F}$ is to find a high-dimensional weight vector $\bm \Omega \in \mathbb{F}$ that minimizes
\begin{equation}
\label{Equ:KRLSRegu}
\min_{\bm \Omega \in \mathbb{F}} \sum_{i=1}^n \|\bm \Omega^T \bm \varphi(\bm u(i))-d(i)\|^2 + \lambda \| \bm \Omega\|^2_\mathbb{F}
\end{equation}
where $\|\cdot\|^2_\mathbb{F}$ denotes the norm in $\mathbb{F}$.
\subsubsection{Approximate linear dependency for sparsification}
Similar to RLS, KRLS achieves high accuracy and has fast convergence rate in stationary scenario. However the good performance is obtained at a  cost of high computation complexity, $O(n^2)$, $n$ is the number of processed sample.

In order to decrease the computational complexity of KRLS, sparisification techniques are adopted. Specifically, sparsification in kernel methods is related to less computational complexity and memory consumption. Furthermore, the system's generalization ability is also influenced by sparsification in machine-learning algorithms and signal processing \cite{kernelbook}. Novelty Criterion, Approximate Linear Dependency (ALD) \cite{KRLS} , Prediction Various, Surprise and Quantized techniques are common strategies for sparisification. Among them, ALD is an effective sparisification technique for KRLS because it is solved in the feature space, unlike most of the other techniques.  Before introducing ALD, several concepts should be interpreted:

Network size: the number of data utilized to describe  the system model.

Center $\bm c$:  the data utilized to build the system. Therefore, the number of centers equals to network size.

Center dictionary $\mathcal{C}$: the set of all centers $\bm c$.

Suppose after having observed $n-1$ training samples, we have established a center dictionary $\mathcal{C}(n-1) = \{\bm{ c}_i\}_{i=1}^{K(n-1)}$, where $K(n-1)$ is the cardinality of $\mathcal{C}(n-1)$. When a new sample $\{\bm u(n), d(n)\}$ is presented, ALD tests whether there exists a coefficient $\bm a(n) =(a_1, \ldots, a_{K(n-1)})$ satisfying
\begin{equation}
d_2 = \min_{\bm a}\left\Vert \sum_{i=1}^{K(n-1)}a_{ni}\bm\varphi(\bm c_i) - \bm\varphi(\bm u(n))\right \Vert^2 \leq \delta
\end{equation}
where $\delta $, called the approximation level, is the threshold determining the sparisification level as well as system accuracy. If this condition is satisfied, the new sample in feature space could be linearly approximated by the centers in $\mathcal{C}(n-1)$. Therefore the effect of this sample to the mapping can be expressed through existing centers in center dictionary and there is no need to augment the center dictionary. Otherwise, the new sample whose feature vector is not approximately dependent on the samples should be added into dictionary and, consequently, a new coefficient corresponding to this center will be included. Through straight calculation it is easy to obtain,
\begin{equation}
\bm a(n) = \bm{\tilde K}(n-1)^{-1} \bm h(n)
\end{equation}
\begin{equation}
d_2 = \kappa(\bm u(n), \bm u(n)) - \bm h(n)^T \bm a(n)
\end{equation}
where $\bm h(n) = [\kappa(\bm c_1, \bm u(n)), \ldots,\kappa(\bm c_{K(n-1)}, \bm u(n))]^T$, the matrix $[\bm{\tilde K}(n-1)]_{ij} = \kappa(\bm c_i,\bm c_j) $. ALD not only is an effective approach to sparsification but also improves the overall stability of the algorithm because of its relation with the eigenvalue of $\bm{\tilde K}$ \cite{kernelbook}.
\subsubsection{Regularized KRLS-ALD}
Now they are two simply way to deal with KRLS-ALD: 1) Setting regularization term equal to 0 \cite{KRLS}. 2) Discarding the samples outside of the center dictionary and ignoring the influence of these samples \cite{kernelbook}. However, both of these two strategies have disadvantages. With the sparsification, the probability of overfitting decreases, even not happens when data number is small enough. This is the inspiration that method 1 used to set the regularization term equal to 0. Unfortunately, the success of this method depends on  the extent of sparsification. If the approximation level is not large enough, it doesn't mitigate overfitting. Owning to discarding useful information, the convergency speed and accuracy performance of the second method may be not satisfactory. In order to overcome these drawbacks, I proposed a general structure of regularized KRLS-ALD.

Define the matrices $\bm \Phi(n) = [\bm\varphi(\bm u(1)),\ldots,\bm\varphi(\bm u(n))]$, $\bm{\tilde{\Phi}}(n) = [\bm\varphi(\bm c_n),\ldots, \bm\varphi(\bm c_{K(n)})]$, according to \cite{KRLS},
\begin{equation}
\bm\Phi(n) = \bm{\tilde{\Phi}}(n) \bm A(n)^T + \bm \Phi(n)^{res}
\end{equation}
\begin{equation}
\bm \Omega(n+1) = \bm{\tilde\Phi}(n)\bm{\tilde \alpha}(n+1)
\end{equation}
where $\bm A(n) =[\bm a(1),...,\bm a(n)]^T$. Then, the cost function becomes
\begin{equation}
\label{Equ:costfunction}
\begin{split}
L(\bm{\tilde \alpha}(n+1)) &=\sum_{i=1}^n \|\bm \Omega(n)^T \bm \varphi(\bm u(i))-d(i)\|^2 + \lambda \| \bm \Omega(n)\|^2_\mathbb{F} \\
&= \|\bm \Phi(n)^T \bm{\tilde\Phi}(n) \bm{\tilde \alpha}(n+1) - \bm d(n)\|^2 + \lambda \|\bm \Omega(n) \|^2_\mathbb{F}  \\
& \approx \|\bm A(n) \bm {\tilde K}(n) \bm{\tilde \alpha}(n+1) - \bm d(n) \|^2 + \lambda\| \bm{\tilde\Phi}(n)\bm{\tilde \alpha}(n+1) \|^2_\mathbb{F} \\
\end{split}
\end{equation}
where $\bm d(n) = [d(1),...d(n)]^T$. In order to minimize $L(\bm{\tilde \alpha}(n))$, we take the derivative with respect to $\bm{\tilde \alpha}(n)$ and obtain
\begin{equation}
\label{Equ:derivative}
\frac{\partial L}{\partial \bm{\tilde \alpha}}(n+1) = 2(\bm A(n) \bm {\tilde K}(n))^T (\bm A(n) \bm {\tilde K}(n) \bm{\tilde \alpha}(n+1) - \bm d(n)) + 2\lambda  \bm{\tilde\Phi}(n)^T \bm{\tilde\Phi}(n)\bm{\tilde \alpha}(n+1)
\end{equation}
At the extremum the system  solution is:
\begin{equation}
\label{Equ:solution}
\bm{\tilde \alpha}(n+1)= [\bm A(n)^T \bm A(n) \bm{\tilde K}(n) + \lambda \bm I]^{-1}\bm A(n)^T \bm d(n)
\end{equation}
In the online scenario, at each time step, we are faced with either one of the following two cases.
\begin{enumerate}
  \item $ \bm \varphi(\bm x(n))$ could be approximated by $\mathcal C(n-1)$ according to ALD, that is $d_2 \leq \delta$. In this case, $\mathcal{C}(n) = \mathcal{C}(n-1)$.
  \item $d_2 > \delta$ and the new data $\bm \varphi(\bm x_n)$ is not ALD on $\mathcal{C}(n-1)$. Therefore, $\mathcal{C}(n) = \mathcal C(n-1) \bigcup \{\bm x(n)\}$.
\end{enumerate}
The key issue in this problem is how to design a iterative solution to obtain $\bm{\tilde \alpha}_n$. In the following, we denote $\bm P(n) = [\bm A(n)^T \bm A(n) \bm {\tilde K}(n)+ \lambda \bm I]^{-1}$ and derive the KRLS with regularization for each of these two cases.

\emph{Dictionary doesn't change:} In this case, $\bm {\tilde \Phi} (n)= \bm {\tilde \Phi}(n-1)$ and hence $\bm {\tilde K}(n) = \bm {\tilde K}(n-1)$. Only $\bm A$ changes between time steps: $\bm A(n)= [\bm A(n-1)^T, \bm a(n)]^T$. Therefore,
\begin{equation}
\bm A(n)^T \bm A(n) = \bm A(n-1)^T \bm A(n-1) + \bm a(n) \bm a(n)^T
\end{equation}
Such  matrix $\bm P(n)$ can be expressed as $\bm P(n) = [\bm P(n-1)^{-1} + \bm a(n) \bm a(n)^T \bm {\tilde K}(n)]^{-1}$. According to the matrix inversion lemma, assume $P(n-1) = A$, $\bm a(n) = B$, $\bm a(n)^T \bm {\tilde K}(n) = C$, $I=D$, yields,
\begin{equation}\label{Equ:PCenterUnchange}
\bm P(n)= \bm P(n-1)- \frac{\bm P(n-1)\bm a_n\cdot \bm a(n)^T \bm {\tilde K}(n) \cdot \bm P(n-1)}{1 + \bm a(n)^T \bm {\tilde K}(n) \cdot \bm  P(n-1)\bm a(n)}
\end{equation}
Defining $\bm q(n)=\bm P(n-1)\bm a_n/(1 + \bm a(n)^T \bm {\tilde K}(n) \cdot \bm  P(n-1)\bm a(n))$ and $\bm s(n)=\bm a(n)^T \bm {\tilde K}(n)$ ,The coefficient vector $\bm{\tilde \alpha}(n+1)$ could then be expressed as
\begin{equation}
\label{Equ:CoeffCenterUnchange}\begin{split}
\bm{\tilde \alpha}(n+1) & = \bm P(n) \bm A(n)^T \bm d(n) \\
& = \!\left[\!\bm P(n-1)\!-\! \frac{\bm P(n-1)\bm a(n)\cdot \bm a(n)^T \bm {\tilde K}(n) \cdot \bm P(n-1)}{1 + \bm a(n)^T \bm {\tilde K}(n) \cdot \bm P(n-1)\bm a(n)}\!\right]\!\! [\bm A(n-1)^T\bm d(n-1) \!+\! \bm a(n) d(n)] \\
& = \bm{\tilde \alpha}(n) + \frac{\bm P(n-1)\bm a(n)[d(n) - \bm a(n)^T \bm {\tilde K}(n) \bm{\tilde \alpha}(n)]}{1 + \bm a(n)^T \bm {\tilde K}(n) \cdot \bm P(n-1)\bm a(n)}\\
& = \bm{\tilde \alpha}(n)- \bm q(n)(d(n)-\bm s(n))\bm{\tilde \alpha}(n)\\
\end{split}
\end{equation}

\emph{The size of center dictionary increases:} In this condition, $\bm {\tilde \Phi}(n) = [\bm {\tilde \Phi}(n) , \bm \varphi(\bm u(n)) ]$. The matrix $\bm A$ changes to
\begin{equation}
\label{Equ:ACenterChange}
\bm A(n)  = \left[
  \begin{array}{cc}
    \bm A(n-1)  & \bm 0 \\
    \bm 0 & 1\\
  \end{array}
\right]
\end{equation}
Therefore,
\begin{equation}
\bm A(n) ^T \bm A(n)  = \left[
  \begin{array}{cc}
    \bm A(n-1) ^T \bm A(n-1) & \bm 0 \\
    \bm 0 & 1\\
  \end{array}
\right]
\end{equation}
\begin{equation}
\bm P(n)  = \left[
  \begin{array}{cc}
    \bm P(n-1) ^{-1} & \bm A(n-1) ^T \bm A(n-1) \bm h(n)  \\
    \bm h(n) ^T & \lambda + \kappa_{nn}\\
  \end{array}
  \right]^{-1}
\end{equation}
where $\bm h(n) = [\kappa(\bm c_1, \bm u(n) ), \ldots,\kappa(\bm c_{k_{n-1}}, \bm u(n))]^T$ and $\kappa_{nn}$ is the simplification of $\kappa(\bm u(n) , \bm u(n))$. Utilize the block matrix inversion identity, we obtain
\begin{equation}
\label{Equ:PCenterChange}
\bm P(n)  = \gamma(n) ^{-1} \left[
  \begin{array}{cc}
    \bm P(n-1) {-1}\gamma(n)  + \bm z_A(n)  \bm z(n) ^T & -\bm z_{A}(n)   \\
    -\bm z(n) ^T  & 1\\
  \end{array}
  \right]
\end{equation}
where $\gamma(n)  = \lambda + \kappa_{nn} - \bm h(n) ^T \bm z_A(n)  $, $\bm z_A(n)  = \bm P(n-1) \bm A(n-1) ^T \bm A(n)  \bm h(n) $, and $\bm z(n)  =\bm  P(n-1) ^T \bm h(n) $. Such the coefficient vector is updated as
\begin{equation}
\label{Equ:CoeffCenterChange}\begin{split}
\bm{\tilde \alpha}(n+1)  & = \bm P(n+1)  \bm A(n) ^T \bm d(n)  \\
& = \gamma(n) ^{-1} \left[
  \begin{array}{cc}
    \bm P(n-1) \gamma(n)  + \bm z_{A}(n)  \bm z(n) ^T & -\bm z_{A}(n)   \\
    -\bm z(n) ^T  & 1\\
  \end{array}
  \right] \left[
  \begin{array}{c}
  \bm A(n-1) ^T \bm d(n-1)  \\
  d_n \\
  \end{array}
  \right]   \\
  & =\left[\begin{array}{c}
  \bm{\tilde \alpha}(n) - \bm z_{A}(n)  \gamma(n) ^{-1} e(n)  \\
  \gamma(n) ^{-1}e(n)
  \end{array} \right]
\end{split}
\end{equation}
where $e(n) = d(n)  - \bm h_n^T \bm{\tilde \alpha }(n) $.

We now have obtained a recursive algorithm to solve the KRLS with regularization, which is referred to as  is described in pseudocode in Algorithm.\ref{algo_KRLS}.
\begin{algorithm}
\caption{Kernel RLS with regularization algorithm} \label{algo_KRLS}
\begin{algorithmic}
\STATE  \emph{\textbf{Initialization}:} Select the threshold $\delta > 0$ and the regularization parameter $\lambda > 0 $,
\STATE $\mathcal{C}(1) = \{\bm u(1)\}$, $P(1) = [\kappa_{11}+\lambda]^{-1}$, $\bm {\tilde K}(1) = \kappa_{11}$,
\STATE $\bm {\tilde K}(1)^{-1} = \kappa_{11}^{-1}$, $\bm A(1) = 1$, $\bm{\tilde \alpha}(1) = \bm P(1) y(1)$
\FOR {$n = 2,3...$}
\STATE Get the new sample: $(\bm u(n), d(n))$;
\STATE Compute $\bm h(n)$
\STATE ALD test: $\bm a(n) = \bm {\tilde K}(n-1)^{-1} \bm h(n)$
\STATE $d_2(n) = \kappa_{nn}-\bm h(n)^T \bm a(n)$
\IF { $d_2(n) \leq \delta^2$ }  
\STATE  $\mathcal{C}(n) = \mathcal{C}(n-1)$, $\bm {\tilde K}(n) = \bm {\tilde K}(n-1)$, $\bm {\tilde K}(n)^{-1} = \bm {\tilde K}(n-1)^{-1}$
\STATE Compute $\bm{\tilde \alpha}(n+1)$  (Equ.\ref{Equ:CoeffCenterUnchange})
\STATE Update $\bm P(n)$ (Equ.\ref{Equ:PCenterUnchange})
\STATE Update $\bm A(n)$, $\bm A(n) = [\bm A(n-1)^T, \bm a(n)]^T$
\ELSE 
\STATE $\mathcal{C}(n) = \mathcal{C}(n-1) \bigcup \{\bm u(n)\}$
\STATE Compute $\bm{\tilde \alpha}(n+1)$ (Equ.\ref{Equ:CoeffCenterChange})
\STATE Update $\bm P(n)$ (Equ.\ref{Equ:PCenterChange})
\STATE Update $\bm {\tilde K}(n)^{-1}$ and $\bm {\tilde K}(n)$
\STATE Update $\bm A(n)$ (Equ.\ref{Equ:ACenterChange})
\ENDIF
\ENDFOR
\STATE return $\bm{\tilde \alpha}(n+1)$, $\mathcal{C}(n)$
\end{algorithmic}
\end{algorithm}
Table \ref{tab:KRLSComplexity} summarizes the computational costs per iteration for KRLS-ALD with and without regularization.
\begin{table}[!t]
    \caption{\label{tab:KRLSComplexity}Computational costs per iteration}
    \centering
    {\small\begin{tabular} {cc}
    \hline
    \hline
    ALD test &  $O(K(n))^2$\\
    \hline
    Update $\bm P(n)$ & $O(K(n))^2$ \\
     \hline
    Update $\bm{\tilde \alpha}(n)$  & $O(K(n))$ \\
     \hline
    \end{tabular}}
\end{table}

\subsection{Kernel Least Mean Square Algorithm}
The KLMS utilizes the gradient descent techniques to search for the optimization solution of KRLS and is the least mean square algorithm (LMS) in RKHS. KLMS is obtained by minimizing the instantaneous cost function:
\begin{equation}\begin{split}
J(n)  & =\frac{1}{2}e(n)^2 \\
& = \frac{1}{2}\|\bm \Omega^T \bm \varphi(\bm u(n))-d(n)\|^2 \\
\end{split}\end{equation}
 Assume the initial condition of weight is $\bm \Omega(0)=0$, then use the LMS algorithm on RKHS, yields
\begin{equation}
\bm \Omega(n+1) = \bm \Omega(n)+\eta e(n)\bm \varphi(\bm u(n))
\end{equation}
However, the dimensionality of $\bm\varphi(.)$ is very high, so an alternative way is needed.
\begin{equation}\begin{split}
\bm \Omega(n+1) & = \bm \Omega(n)+ \eta e(n)\bm \varphi(\bm u(n)) \\
= & \bm \Omega(n-1) + \eta e(n-1)\bm \varphi(\bm u(n-1))+ \eta e(n)\bm \varphi(\bm u(n))\\
& \ldots \\
& = \eta \sum^{n}_{i=1}e(i)\bm \varphi(\bm u(i))\\
\end{split}\end{equation}
According to the ``kernel trick'' the output of system of the new input $\bm u(n)$ can be expressed as
\begin{equation}\begin{split}
\bm \Omega(n)^T \bm \varphi(\bm u(n)) & = [\eta \sum^{n-1}_{i=1}e(i)\bm \varphi(\bm u(i))]\bm \varphi(\bm u(n))\\
& =\eta \sum^{n-1}_{i=1}e(i)\kappa(\bm u(i),\bm u(n))\\
\end{split}\end{equation}
In conclusion, the learning rule of KLMS in the original space is as show in Algorithm. \ref{Kernel least mean square algorithm}.
\begin{algorithm}
\caption{Kernel least mean square algorithm}
\label{Kernel least mean square algorithm}
\begin{algorithmic}
 \STATE  \emph{\textbf{Initialization}:} Select the stepsize $\eta$,
 \STATE  $ e(1)= d(1) $;
 \STATE  $ y(1) = \eta e(1)$;
 \STATE  \emph{Computation}
 \WHILE {$\{\bm{u}(n), d(n)\}$ \emph{available}}
 \STATE $ y(n) =\eta\sum_{i=1}^{n-1}[e(i)\kappa(\bm{u}(i),\bm{u}(n))]$;
 \STATE $e(n) = d(n) - y(n)$;
 \ENDWHILE
\end{algorithmic}
\end{algorithm}
Among all of the kernel adaptive filters, KLMS is unique. It provides well-posedness solution with finite data \cite{klms} and naturally creates a growing radial-basis function (RBF) network \cite{kernelbook}. Moreover, as an online learning algorithm, KLMS is much simpler to implement, with respect to computational complexity and memory storage, than other batch-model kernel methods. 
\subsection{Kernel Affine Projection Algorithm} The KLMS algorithm simply uses the current values to search the optimal solution. While the affine projection algorithm (APA) adopts better approximation by using the $K$ most recent samples and provides a trade-off between computation complexity and system performance. More interestingly, the KAPA provides a general framework for several exiting techniques, including KLMS, sliding window KRLS and kernel regularization networks \cite{kernelbook}.

\section{Discussion}
Even though kernel adaptive filters have advantages mentioned above and have been proved to be useful in complicated nonlinear regression and classification problems, they still have some drawbacks. For example, the conventional cost function, mean-square error (MSE) criterion, can not obtain the best performance in non-Gaussian situations. Moreover, the linear growing structure with each new sample leads to high computation burden and memory requirement. Therefore, computational complexity and memory increase linearly with time as more samples are processed, particulary for continuous scenarios which hinder so far online application, such as in DSP and FPGA implementations. Therefore to apply these powerful kernel adaptive filters in practice, we need to address two main issues first:
\begin{itemize}
  \item \textbf{How to obtain a robust performance in non-Gaussian situations?}
  \item \textbf{How to decrease the growing computational burden and memory requirement?}
\end{itemize}
These two issues are important aspects to judge system performance and are highly related. Improving robust performance normally results a the computational complexity increase. Therefore, appropriate processes, such as reducing the network size, to decrease computational complexity are necessary. On the other hand, approximation techniques to decrease computation complexity may suffer from system performance getting worse and could be compensated by improving system robust performance. Throughout the literature, similar problems have been studied from different perspectives leading to a myriad of techniques. Yet few techniques have been adopted to kernel adaptive filters. Our goal is to provide appropriate methods which take into account the intrinsic properties of kernel adaptive filters to solve these two problems.

There are various factors influencing kernel adaptive filters performance, such as cost function and kernel function. First, let us consider the cost function. MSE criterion is equivalent to maximum likelihood technique in linear and Gaussian condition  and obtains good performance, while it is not enough in other scenarios. The least mean p - power error (MPE) criterion, builds  a linear weighted combination of various powers of the prediction error, is investigated to solve this problem. However, many free parameters and prior knowledge requirements limit its wide application in practice. Recently, Information theoretic learning (ITL) has been proved more efficient to train adaptive systems. Different from conventional error cost function, ITL optimizes the information content of the prediction error to achieve the best performance in terms of information filtering. By taking into account the whole signal distribution, adaptive systems training through information theoretic criteria have better performance in various applications especially for which Gaussianity assumption is not valid. Furthermore, the nonparametric property in cost function and clear physical meaning also motivate us to introduce the ITL into the kernel adaptive filters to achieve a robust system and propose the kernel maximum correntropy algorithm (KMC).

Even though the accuracy of a system is improved, the redundant computational and memory burden is what we have to face with. Therefore the other important problem for the practical application of kernel adaptive filters is compacting network structure, which helps us decrease the total computational complexity. By including the important information and discarding relatively less useful data, growing and pruning techniques maintain the system network size into an acceptable range. Different from previous sparsification techniques, the quantized kernel least mean square utilized ``redundant'' data to update the network not purely discarded them to obtain a compact system with higher accuracy. As a complement of
spasification, pruning strategies makes the system structure more compact. Sensitive analysis based pruning strategies, discarding the units which make insignificant contribution to the overall network, are robust and widely utilized in various areas. Generally, omitting any information brings the accuracy down. Hence, the growing and pruning techniques is a trade off between system performance and compact structure.
Therefore, a measure of \emph{significance} is adopted in kernel adaptive filter to estimate the influence of process data and decide what information will be discarded.

The significance measure guides the system to fix the network size on a predefined threshold. However, this is not a real nonstationary learning. What we expected is that the network size should optimally be dictated by the complexity of the true system and the signal, while the system accuracy is acceptable. If we solve this problem, we have a truly online algorithm to handle real world nonstationary problems. This problem is transferred as how to obtain a trade-off between system complexity and accuracy performance. A variety of information criteria have been proposed so far to deal with this compromise problem. Among them, Minimal Description Length (MDL) has  great advantages of small computational costs and robust against noise, which have been proven in many applications particular in information learning area. MDL criterion has two formulations: batch model and online model.  Taking the approximation level selection in KRLS-ALD as an example,  the batch model MDL in kernel adaptive filters is illustrated firstly. Then we proposed an  KLMS sparsification algorithm  to explain the online model MDL. Owning to this proposed algorithm  separating the input (feature) space  with quantization techniques, it is called QKLMS-MDL.

This article mainly focus on the improvement of KLMS, the simplest of the kernel adaptive filters, but we believe, further extentions to other kernel adaptive filters, such as kernel recursive least square algorithm (KRLS) will broaden the the scope of the applications currently addressed in KLMS.

\bibliographystyle{unsrt}
\bibliography{ee512}

\end{document}